\pdfoutput=1

\documentclass[11pt]{article}

\usepackage{acl}

\usepackage{times}
\usepackage{latexsym}
\usepackage{multirow}
\usepackage{multicol}
\usepackage{booktabs}
\usepackage{graphicx}
\usepackage{times}
\usepackage{latexsym}
\usepackage{tcolorbox}
\usepackage{xcolor}  
\usepackage{diagbox}
\usepackage[T1]{fontenc}

\usepackage[utf8]{inputenc}

\usepackage{microtype}

\title{PPTC Benchmark: Evaluating Large Language Models for PowerPoint Task Completion}
\author{
Yiduo Guo$^{1}$\footnotemark[1],~~Zekai Zhang$^{1}$\footnotemark[1],~~Yaobo Liang$^{2}$,~~Dongyan Zhao$^{1}$,~~Nan Duan$^{2}$\\ 
$^1$Peking University\\
$^2$Microsoft Research Asia\\
\texttt{yiduo@stu.pku.edu.cn,justinzzk@stu.pku.edu.cn,yaobo.liang@microsoft.com}\\\texttt{zhaody@pku.edu.cn,nanduan@microsoft.com}\\
}
\begin{document}
\maketitle
\renewcommand{\thefootnote}{\fnsymbol{footnote}}
\footnotetext[1]{Equal contribution}
\begin{abstract}
Recent evaluations of Large Language Models (LLMs) have centered around testing their zero-shot/few-shot capabilities for basic natural language tasks and their ability to translate instructions into tool APIs. However, the evaluation of LLMs utilizing complex tools to finish multi-turn, multi-modal instructions in a complex multi-modal environment has not been investigated. To address this gap, we introduce the PowerPoint Task Completion (PPTC) benchmark to assess LLMs' ability to create and edit PPT files based on user instructions. It contains 279 multi-turn sessions covering diverse topics and hundreds of instructions involving multi-modal operations. We also propose the PPTX-Match Evaluation System that evaluates if LLMs finish the instruction based on the prediction file rather than the label API sequence, thus it supports various LLM-generated API sequences. We measure 3 closed LLMs and 6 open-source LLMs. The results show that GPT-4 outperforms other LLMs with 75.1\% accuracy in single-turn dialogue testing but faces challenges in completing entire sessions, achieving just 6\% session accuracy. We find three main error causes in our benchmark: error accumulation in the multi-turn session,
long PPT template processing, and multi-modality perception. These pose great challenges for future LLM and agent systems. We release the data, code, and evaluation system of PPTC at \url{https://github.com/gydpku/PPTC}.
\begin{figure}[t]
\centering 
\vspace{-2mm}

\includegraphics[height=0.525\textwidth,width=0.485\textwidth]{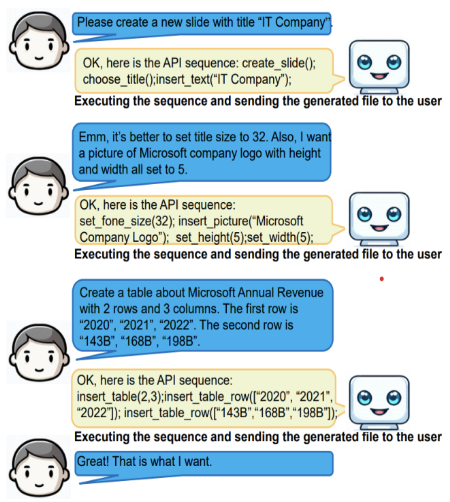} 
\vspace{-3mm}
\caption{Within our benchmark, we simulate this multi-turn dialogue scenario between humans and LLMs to evaluate LLMs' PPT task completion performance.}
\label{Fig.example}
\vspace{-5mm}
\end{figure}
\end{abstract}

\section{Introduction}
\begin{figure*}[t]
\centering 
\vspace{-2mm}

\includegraphics[height=0.7\textwidth,width=1\textwidth]{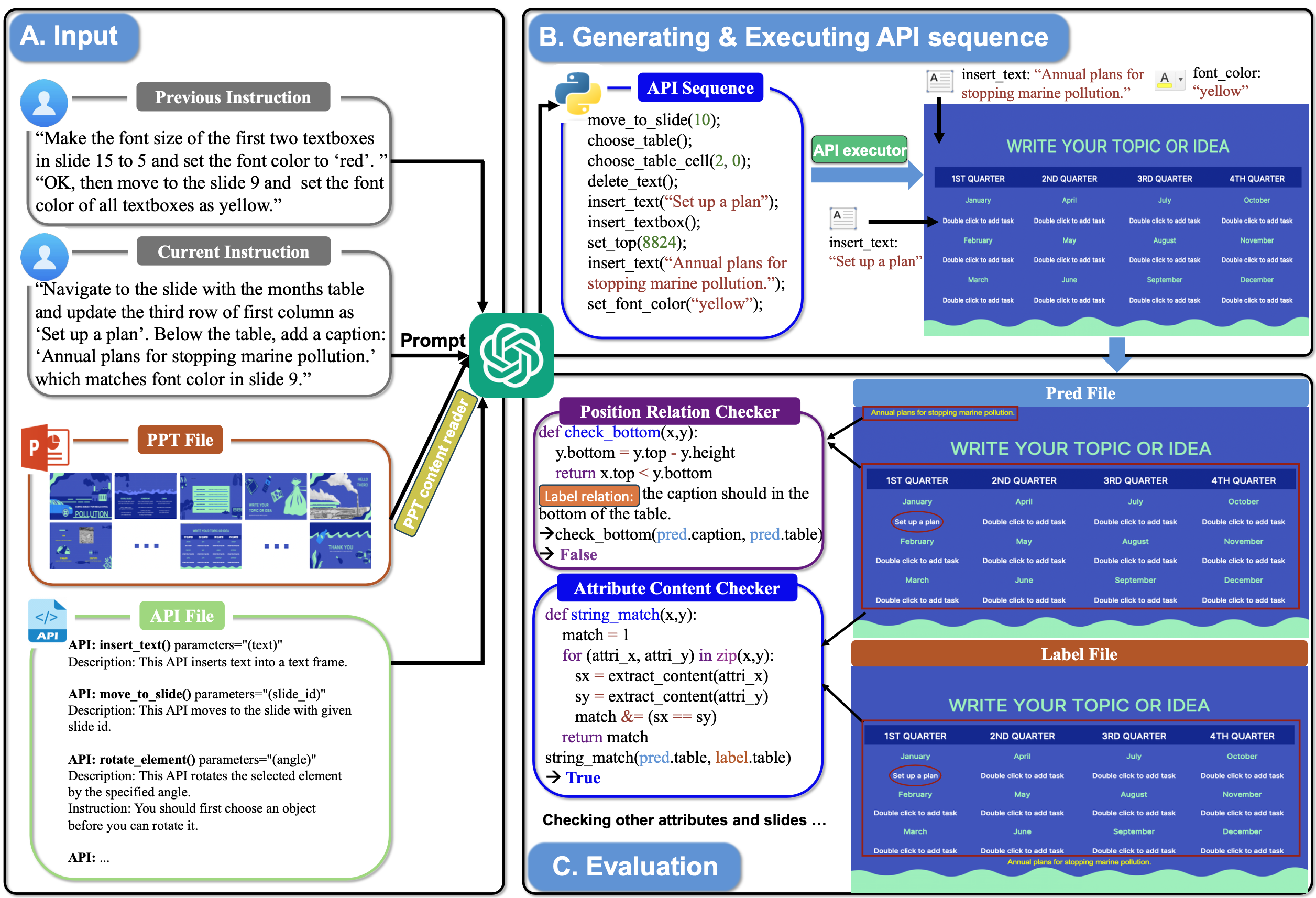} 
\vspace{-3mm}
\caption{We illustrate how LLMs complete one turn in a session. (A) To prompt the LLM, we provide it with the current instruction, previous instructions (dialogue history), PPT file content, and the API reference file. 'PPT reader' is a function that transforms the PPT file into the text-based format as the PPT file content. (B) The LLM then generates the API sequence and executes it to obtain the prediction PPT file. (C) We evaluate attributes and position relations in the prediction file.}
\label{Fig.completion}
\vspace{-5mm}
\end{figure*}
Recent evaluation works for Large Language Models (e.g.~ChatGPT and GPT-4~\cite{openai2023gpt4}) focus on their zero-shot/few-shot abilities on basic natural language tasks~\cite{jiao2023chatgpt,zhong2023agieval,wang2023chatgpt,qin2023chatgpt} and their tool-use ability to generate APIs for solving user instructions, such as basic APIs like a calculator in tool transformer~\cite{schick2023toolformer}, RapidAPIs in ToolLLM~\cite{qin2023toolllm}, and hugggingface APIs in Gorilla~\cite{patil2023gorilla}. However, these tool-use works emphasize the translation of natural language instructions into APIs and ignore the challenge of using APIs in the observation of complex multi-modal environments to finish user instructions.
Also, their evaluation approach focuses on comparing the generated APIs with the label API sequence, assuming there's only one unique solution. This approach becomes impracticable in situations with multiple/unlimited correct solutions. To address these challenges, we introduce
\textbf{P}ower-\textbf{P}oint \textbf{T}ask \textbf{C}ompletion (PPTC), a benchmark that measures LLMs' performance in creating and editing PPT file tasks based on user instructions. We choose PowerPoint as it includes various elements like textbox, table, and image and supports a wider range of APIs than Word and Excel. 

Our benchmark has three distinctive features from other task completion benchmarks:~(1) \textbf{Multi-turn dialogue with varying difficulty}. Our PPTC benchmark simulates the multi-turn dialogue session between the user and the LLM (see Figure~\ref{Fig.example}) and contains 279 multi-turn sessions. Each multi-turn session in our benchmark includes 2 to 17 turns. Each turn consists of a user instruction that describes the user's needs, a feasible solution that provides the correct solution, and the resulting label output file. Some turns can be easily addressed using a single API, while over half of the instructions require multiple APIs for completion. We provide the LLM with a reference API file that contains all feasible APIs for selection. (2) \textbf{Multi-modality}. Finishing the instruction of our benchmark requires understanding the multi-modal PPT file content and using multi-modal API operations (e.g., PPTC has 268 image operation-related instructions). (3) \textbf{Evaluation based on the final status}: We propose the PPTX-Match Evaluation system to evaluate the LLM's outcome. To identify if the LLM completes the instruction, it checks the PPT file produced by executing the LLM-generated APIs rather than the LLM-generated APIs, thus all API sequences that lead to the correct final status are acceptable. 

To finish the instruction, we use the current instruction, past turns' instructions (dialogue history), the PPT file content (specific environment information), and the reference API file as the input to prompt the LLM to generate an API sequence as the solution (See Figure~\ref{Fig.completion} (A)). Then we use the API executor to execute the API sequence and return the user the resulting PPT file~(See Figure~\ref{Fig.completion} (B)). We name the resulting PPT file as the prediction file. In the evaluation step~(See Figure~\ref{Fig.completion} (C)), the PPTX-Match Evaluation system first uses the Python-PPTX library to extract all attributes from the prediction PPT file and the label output file. Then it uses the position relation checker to check if objects' positions conform to the label relation and the attribute content checker to check if the attribute's content is matched with the corresponding label attribute's content. The LLM correctly completes the current turn's instruction if all attributes of the file pass these tests. Evaluation metrics include turn-based accuracy which is the ratio of correctly completed turns to the total number of turns and session-based accuracy which is the ratio of correctly completed sessions to the overall session count.
We measure the performance of three closed-source LLMs (GPT-4, ChatGPT, and Davince-003) and six open-source LLMs (e.g., LLaMa-2) in our benchmark. We further test planning (e.g., CoT~\cite{wei2022chain}) and content selection algorithms' performance based on GPT-4.
Experiment results show that GPT-4 is the strongest LLM among all LLMs but still encounters challenges when completing entire multi-turn sessions. For example, although GPT-4 achieves 75.1\% turn-based accuracy in the creating new PPT file task, it only achieves 22.7\% session-based accuracy as errors made in previous turns. GPT-4 and other LLMs also struggle to process long PPT templates (complex file environment). For example, GPT-4 only achieves 38.1\% turn-based accuracy in the editing task. We further find that GPT-4 struggles to finish instructions involving non-text modality operations, especially for position-related operations, such as 'Put object A on the top of the slide'. It only achieves 24\% accuracy in these instructions. 

In summary, this paper has the following contributions:  

(1) We propose the PowerPoint Task Completion benchmark to measure LLM's task completion performance within the PowerPoint official software. This benchmark contains 279 multi-turn sessions with hundreds of multi-modal instructions in the complex multi-modal environment. 

(2) We propose the PPTX-evaluation system to automatically measure LLMs' performance in our benchmark. We test 3 closed-source LLMs and 6 open-source LLMs and find that GPT-4 is the strongest LLM among all LLMs. 

(3) We further analyze LLMs in our benchmarks and find three key error factors: error accumulation in the session, long PPT template processing, and multi-modality perception. These findings pose significant challenges for future LLMs and LLM-based systems.

\section{PPTC Benchmark}
In this section, we introduce our Power-Point Task Completion (PPTC) benchmark, including the overview of our benchmark, its collection and validation process, and the PPTX-Match Evaluation System for evaluation. We further analyze the statistics information of our benchmark.

\subsection{Benchmark Overview}
\textbf{Benchmark components}
Our benchmark focuses on two basic tasks within PowerPoint: creating the new PPT file and editing the existing long PPT template for measuring long PPT Content understanding. We have gathered 229 multi-turn dialogue sessions for creating the new PPT file and 50 sessions for editing existing templates. Each multi-turn session includes 2 to 17 turns. Each turn comprises three parts: (1)  the user instruction (2) the label output file as the ground truth (3) one feasible API sequence for finishing the instruction. Our benchmark also contains an API reference file that includes 49 feasible APIs for various operations and can complete all instructions in our benchmark. For each API, we describe its functionality and arguments and provide usage guidelines. For complex APIs, we also offer example cases. We list the details of all APIs in Appendix~\ref{appendix:apis}.

\textbf{Task description} To complete the instruction in one turn, in general, the AI assistant must comprehend the user's current and prior instructions for context. It should also analyze the content of the PPT file to identify relevant objects mentioned in the instruction. Additionally, it needs to select appropriate APIs from a reference API file to achieve the user's goals. So we use these as the input of the AI assistant and it should output an API sequence as the solution. Then, it executes this API sequence and provides the user with the resulting PPT file as its response (See the whole process in Figure~\ref{Fig.completion}). 

\textbf{Addressing LLM limitations in our benchmark}
Compared to the general AI assistant, LLMs still have two limitations for completing the task in our benchmarks: (1)  LLMs can not directly process the PPT file. So we provide a PPT reader function that extracts all shapes and their information from the PPT file and transforms them into the text format as the PPT file content. Then LLMs can understand and process the PPT file content. (2) LLMs cannot directly use PPT software through a keyboard and mouse. Therefore, we have defined PPT APIs based on the operational logic within the PPT software. and provide an implementation for these APIs in Python that can swiftly generate PPT files. In future work, it may be possible to explore the use of large multimodal models to understand on-screen content and implement APIs using a keyboard and mouse.

\subsection{Benchmark Collection}
\textbf{Design Principles}
We follow these principles to design our  benchmark:
(1) Multi-turn instructions: One session in our benchmark should contain multi-turn instructions to finish the user's complex need.
(2) Instructions of varying difficulty: Some instructions can be achieved with a single API, while others necessitate a sequence of APIs for successful completion.
(3) Diverse multimodal operations: User instructions should cover a wide range of operations on PPT, such as text-related, image-related, and position-related APIs.
{\color{black}(4) Topic Consistency: The dialogue in a session should center around the session topic. Each user instruction in a session aligns closely with the previous instructions (the context), ensuring a coherent and contextually relevant dialogue flow. 
(5) Practicability First: The session topic and specific instructions should simulate the user's need in real world
}



\textbf{Benchmark Collection and Validation}
To collect user instructions, we engage 6 skilled crowd workers who craft instructions in accordance with the principles we've outlined. Our crowd workers comprise professional data science engineers well-versed in PowerPoint. To achieve practicability first, we request crowd workers to write instructions based on their actual PowerPoint experience. On each session, the workers are asked to first find and list a practicable session topic. For the editing PPT template task, the topic must based on the template file background and is practicable to the template\footnote{We collect 50 PPT templates from the SlidesCarnival website (\url{https://www.slidescarnival.com/}). SlidesCarnival is a free and open-source PPT template website. Each session in the editing task has a unique template. We encourage topic diversity in our templates. We remove templates that are too short (2$\sim$5 slides) and have repeated topics.}. To achieve multi-instructions and topic consistency, the workers write instructions step by step and make them consistent with the topic. To achieve diverse multi-operations, we ask them not to write session that only involves a single modality operation. Each worker takes on a specific role in the instructional writing work and is encouraged to write instructions in his/her own words. Workers were asked to spend at least 20 minutes on every session. We delete repeated sessions and short sessions that have no more than 50 tokens.

Then we ask the seventh worker to write the feasible API sequence with minimal API usage for each instruction. Next, the workers create the PPT label file by using the provided API sequence.  
During the whole process, the principal engineer reviews and refines the instructions and API sequences written by the above 7 workers for initial quality assurance. 

To ensure the data quality of this benchmark, the three authors of this paper further undertake the following validation steps:
(1) Assessing Instruction Clarity and Relevance: They examine whether the instructions are clear, contextually related to the session topic, and align with the ongoing conversation.
(2) API Sequence Execution: The authors execute the provided API sequences to identify and rectify coding errors.
(3) Goal Achievement Check: They verify if the instructions' intended goals are successfully completed in the label files.

In the event that errors are identified during this validation process, the authors promptly report them to the respective workers for revision.  The three authors are computer science senior students and researchers.

\begin{figure*}[t]
\centering 
\vspace{-2mm}

\includegraphics[height=0.325\textwidth,width=1.01\textwidth]{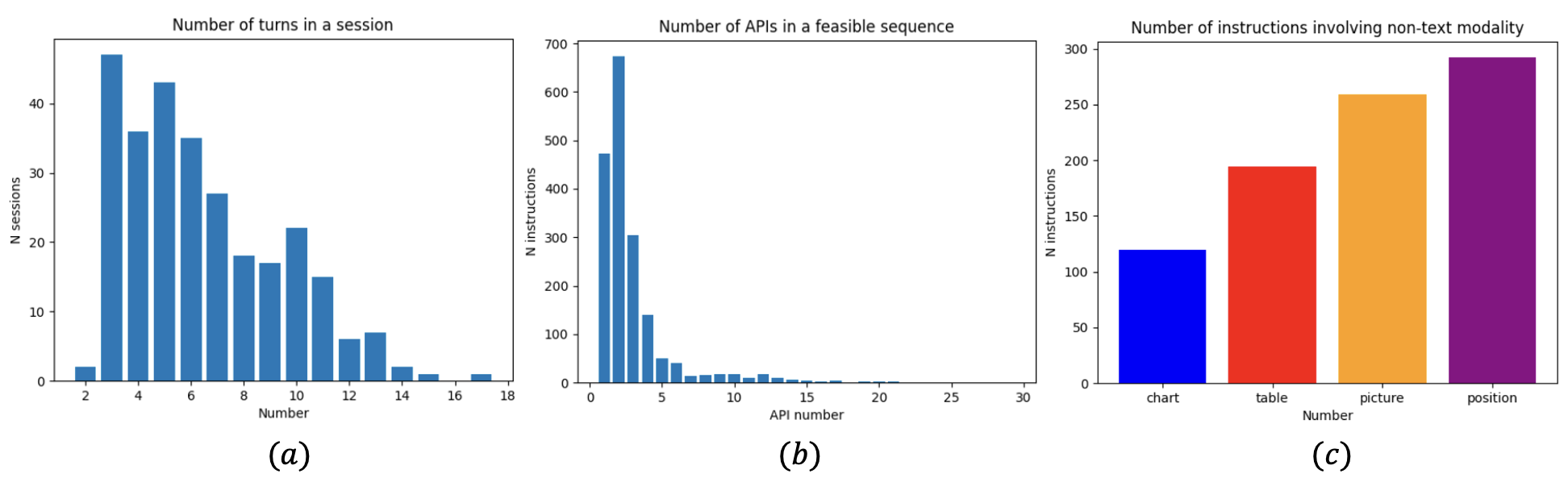} 
\vspace{-3mm}
\caption{Statistics for PPTC. a) Session turn number distribution. b) Instruction API number distribution (tokens). c) Distribution of instructions involving Chart, Table, Picture, and Position. Instructions involving 'Position' need the system to conduct position-related operations based on the understanding of spatial information. Note that one instruction may involve multiple different modalities.}
\label{Fig.statistics}
\vspace{-4mm}
\end{figure*}
\subsection{PPTX-Match Evaluation System}
We design the PPTX-Match Evaluation System to evaluate LLMs' performance on the PPTC benchmark. Specifically, our PPTX-Match Evaluation System first uses a Python-PPTX Content Reader Module to iterate over all shapes in the prediction PPT file produced with the LLM and the label output file. A shape in the PPTX library typically refers to an individual object, such as a text box or table. Then our system extracts attributes like text, style, and position of the shapes using the PPTX library.
Next, we check all attributes from the prediction PPT file. For non-position attributes (e.g., text content), we first convert it and the corresponding attribute in the label PPT file into two strings, and then we use the Exact Match method to examine if the two strings are the same. If they are different or we do not find the corresponding attribute in the label file, then we find an incorrect match. For the position attribute (e.g., location information), we focus on checking if the objects in the prediction PPT file satisfy the correct position relation <A, B, REL>, where A and B are objects that should satisfy the relation REL. In the  benchmark collection process, we ask crowd workers to label the position relation that objects should satisfy to finish the instruction. During the evaluation phase, we extract the position attributes of objects A and B and use predefined functions to verify if the objects' position attributes satisfy the position relation.

If there are no incorrect matches for all non-position attributes and no rule violations for all position-related attributes, we consider the LLM has successfully completed the user instruction.

\subsection{Benchmark Statistics Analysis}
{\color{ black}To understand the properties of PPTC, we analyze
the instructions and APIs in the benchmark.
Specifically, we explore (i) the number of turns in a session, (ii) the difficulty of the instruction in terms of the number
of APIs required to finish it, and (iii) the
number of multi-modality instructions.
We report statistics about the PPTC benchmark in Figure~\ref{Fig.statistics}.

\textbf{The number of turns in a session} The session turn number distribution~(Figure~\ref{Fig.statistics} (a)), measured
as the number of turns in a session, shows that all sessions in our benchmark have at least two turns and almost all sessions have at least 3 turns (between 3
and 13 turns for the 5th to 95th percentile, respectively). The longest session has 17 turns, which is very challenging as the errors made in previous turns can influence the completion of the current instruction. 

\textbf{Diffculty varies in APIs number} The number of APIs in a sequence falls between 1 and
5 for the 5th to 95th percentile~(Figure~\ref{Fig.statistics} (b)), respectively, shows that our instructions' difficulty varies from a simple sentence that can be finished by one API to a complex instruction that requires the LLM to generate multiple APIs. The longest API sequence consists of 29 APIs. Generating long API sequences is very challenging as the LLM needs to understand sub-goals in the complex instruction, select appropriate APIs from the file, and generate APIs in a reliable order.

\textbf{Rich multi-modal instructions} Our benchmark has hundreds of instructions that involve multi-modalities content~(Figure~\ref{Fig.statistics} (c)). The "chart" modality has the fewest instructions, with 120, while the "position" modality has the most, with 292 instructions. To finish these instructions, LLMs need to employ related-modal APIs based on the understanding of multi-modal file content. }

\section{Algorithms}
In this section, we introduce the algorithms we considered to enhance the LLM's performance in our benchmark. 
These algorithms can be categorized into two approaches: planning algorithms that help the LLM in decomposing the user instruction and solving it step by step and selection algorithms that assist the LLM in choosing important environmental information or APIs.

\subsection{Planning Algorithms}

Complex user instructions often require multiple intermediate steps to complete. We mainly consider two planning algorithms:

\textbf{Zero-shot-CoT }\cite{kojima2022large} enables LLMs to autonomously generate intermediate reasoning processes for complex
instruction by prompting LLMs to "Let's think step by step".

\textbf{Tree of Thoughts (ToT) }\cite{yao2023tree} enables LLMs to follow tree-like reasoning paths, where each tree node represents a thinking state. It leverages LLMs to generate evaluations or votes on different thoughts. 

\subsection{Selection Algorithms}

{\color{ black}Combining the whole PPT file and the whole API file into the LLM's input can result in an overwhelming amount of redundant information, such as irrelevant file content and unhelpful APIs. Filtering the redundant information would improve the efficiency of the LLM.}
In this context, we primarily focus on two algorithms for selecting the PPT file content and APIs, respectively: 

\textbf{Content Selection algorithm} Firstly, we extract all shapes of the PPT file by Python-PPTX. Next, we prompt the LLM to select the shapes for completing the user's instruction. We show the prompt in Figure~\ref{fig.content_selection}, in which we add three demonstration examples to guide the LLM to do selection. In this algorithm, we replace the whole PPT file with the selected shapes when prompting the LLM to generate the API sequence.  

\textbf{API Selection algorithm} The API selection algorithm is based on the embedding similarity to select the most relevant APIs for user instructions. Specifically, we use the text embedding API to get the embeddings of all API descriptions and the current user instruction. Next, we compute the cosine similarity between the instruction embedding and each API description's embedding and rank them based on the similarity score. In this algorithm, we replace the whole reference API file with the top $k$ APIs when prompting the LLM to generate the API sequence. 

\section{Experiments}

\subsection{Large Language Models Selected for Evaluation}
Here, we assess different cutting-edge large language models using our benchmark. These chosen models showcase a wide array of capabilities and are highly regarded in the field. The evaluated large language models include 3 closed-source LLMs and 6 open-source LLMs:

\begin{itemize}
\item \textbf{\textit{GPT-4}}~\cite{openai2023gpt4}: The latest LLM in the GPT series. GPT-4 is a cutting-edge, large-scale generative pre-trained transformer model. It offers improved performance and a wider knowledge base compared to its predecessors. It showcases human-level proficiency in several scenarios. 
\item \textbf{\textit{ChatGPT}}: ChatGPT is a conversational AI model crafted for dynamic interactions. It's learned from extensive instruction data and fine-tuned through reinforcement learning with human feedback (RLHF). This empowers it to deliver responses that align with human expectations, maintaining context and coherence in conversations.
\item \textbf{\textit{Text-Davinci-003}}~\cite{brown2020language}: GPT-3.5 sits between GPT-3 and GPT-4, enhancing performance via additional instruction tuning. It acts as a link between these models, facilitating comparison. We've chosen the \textbf{Text-Davinci-003} variant from the GPT-3.5 series for our evaluation.

\item \textbf{\textit{LLaMa-2-Chat}}~\cite{touvron2023llama}: LLaMa 2, an auto-regressive open-source language model, employs an optimized transformer design. Chat versions utilize supervised fine-tuning (SFT) and reinforcement learning with human feedback (RLHF) to match human preferences for helpfulness and safety.
\item \textbf{\textit{Baichuan-Chat}}: It is a transformer model trained on approximately 1.2 trillion tokens. It supports both Chinese and English, with a context window length of 4096. 

\item \textbf{\textit{Baichuan-2-Chat}}~\cite{yang2023baichuan}: It is a large-scale multilingual language model trained from scratch, on 2.6 trillion tokens. The chat version uses Supervised Fine-Tuning
(SFT) and Reinforcement Learning from Human Feedback (RLHF) to align with humans.

\item \textbf{\textit{WizardLM v1.2}}~\cite{xu2023wizardlm}: WizardLM v1.2 is finetuned from LLaMa 2 using supervised instruction fine-tuning, where instructions are created by rewriting the initial instructions step by step.
\item \textbf{\textit{Vicuna v1.5 (16k)}}~\cite{chiang2023vicuna}: Vicuna v1.5 (16k) is finetuned from LLaMa 2 using supervised instruction fine-tuning and linear RoPE scaling. It's trained on about 125K conversations sourced from ShareGPT.com.
\item \textbf{\textit{Code-LLaMa-instruct}}~\cite{chiang2023vicuna}: Code-LLaMa is a LLaMa-based LLM designed for general code completion and understanding. Its instruction version further supports the chat function with users. Code LLaMa models feature a multitask training objective consisting of both autoregressive and causal infilling prediction (predicting the missing part of a program given a surrounding context).
\end{itemize}
\begin{figure}
\tcbset{colback=green!-15!white, colframe=green!65! black, fonttitle=\bfseries}
\begin{tcolorbox}[title=Inference prompt in PPTC, sidebyside align=top]
\small{(\textbf{Task instruction}) You are an AI assistant to help the user to operate PowerPoint and edit the contents.\\}
\small{ Give you the user instruction:<Current user instruction>, you can complete it based on the following APIs and PPT file content. Current you are at page <Page id>. Please finish the user instruction with the functions you have.
Don't generate instructions beyond what the user has instructed. 
Don't guess what the user may instruct in the next step and generete API for them.
Don't use python loop to call API. You can only call API once in one line.
If the user does not specify the page to be modified, you can directly start using the APIs without having to navigate to other pages.

You need to generate code which can finish the user instruction. The multiple lines of code should be surrounded by <code> and </code> such as:
<code>
API();
API();
</code>

For example, if the user instruction is "create a slide", then the answer should be:
\\<code>
create\_slide();
</code>}
\\\\\small{(\textbf{API file}) Now, you have access to a list of PowerPoint APIs with the following functions: <APIs and their descriptions> \\(e.g.,API(name="set$\_$width", parameters="(width)", \\description="This API sets the width of the selected object.",
        \\parameter$\_$description="It takes one parameter 'width', the width of an object in centimeters as float.",
        \\composition$\_$instruction="You should first choose an object before you can change the width of it.",\\api$\_$desc="width of picture and shapes")
)
\\\\(\textbf{PPT file content}) All the PPT contents are:
\\<Begin of PPT>
\\\textit{Turn-based: <Parsed PPT file content of the label PPT file of the previous turns>\\Session-based: <Parsed PPT file content of the LLM prediction file of the previous turns>}\\<End of PPT>
\\\\\small{(\textbf{Dialogue history}) 
\\¬User¬:
Hello!
\\¬AI¬:
Hi there! How can I help you?
\\¬User¬:
<the first instruction>
\\¬AI¬:
\\\textit{Turn-based: <the correct feasible API sequence>,
\\Session-based: <the LLM-generated API sequence>}
\\...
\\¬User¬:
<Current user instruction>. Surrounding your answer with <code> and </code>.
\\¬AI¬:}}
\end{tcolorbox}
\captionof{figure}{The inference prompt we used in both turn-based and session-based evaluation settings. In the turn-based evaluation, we assess the LLM's performance for the current turn and assume the LLM has correctly finished previous turns. We then use feasible API sequences of previous turns as the AI response in the dialogue history and parse the label file of previous turns as the PPT file content. In the session-based evaluation, we evaluate the completion of the entire session and do not assume the LLM has correctly finished previous turns. We use the LLM's generated API sequences as the response and parsed the LLM prediction file as the PPT file content.}
\label{inference_prompt}
\end{figure}
\subsection{Experimental Setup}
In this section, we provide an overview of the experimental setup utilized to assess the performance of LLMs on our PPTC benchmark. 

\subsubsection{Turn-Based and Session-Based Evaluations}
We consider two performance evaluation approaches in our benchmark: turn-based and session-based evaluations. For the turn-based evaluation, we measure the LLM's ability to finish a single turn. Specifically, in this evaluation, we assume that the previous turns have been correctly finished, and we prompt the LLM to generate the API sequence to finish the current turn's user instruction. The prompt consists of the task instruction for finishing the current user instruction, the API file containing feasible APIs, the parsed PPT file content from the PPT file, and dialogue history consisting of instructions of previous turns with their feasible API sequences (see the left of Figure~\ref{inference_prompt}). 
For the session-based evaluation, we measure the LLM's ability to finish a session containing multiple turns. For all turns in a session, we prompt the LLM to finish them sequentially. The prompt in this evaluation has two differences: the API solutions for previous turns in dialogue history are the outputs of the LLM instead of the correct API sequences. (2) The PPT content is parsed from the PPT file obtained by executing the previous outputs of the LLM (see the right of Figure~\ref{inference_prompt}). That means the error made by LLMs in previous turns would influence subsequent turns.

\textbf{Metrics} For turn-based evaluation, we report the turn-based accuracy as the ratio of the number of successfully finished turns to the total number of turns. We also report the average token number of the input of one turn and the average API number for finishing one turn as the cost measurement. For session-based evaluation, we report the session-based accuracy as the ratio of the number of successfully finished sessions to the total number of sessions. We also report the average value of the token number of all inputs in one session and the average API number required to complete one session as the cost measurement. 

\subsection{Implementation Details}
All experiments were conducted using the respective language models' API provided by Azure OpenAI Service\footnote{\url{https://azure.microsoft.com/en-us/products/cognitive-services/openai-service}}. Azure OpenAI services offer two API types: completion and chat completion. Completion API generates text from prompts, while chat completion API responds based on conversation history and new input. We use the completion API for Text-Davinci-003 and the chat completion API for ChatGPT and GPT-4. We set a temperature of zero for deterministic output and a max token limit of 2048. The frequency penalty and top p are kept at their default values of zero and 1, respectively. We use the text-embedding-ada-002 API as the embedding API in the API selection algorithm and set $k$ as 15. For open-source LLMs, we choose the chat version of LLaMa-2, the v1.2 version of WizardLM, and the chat version of Baichuan as our open-source LLMs. We choose the 13 billion parameters model of the three LLMs. 

For the zero-shot CoT method, we add the sentence 'Let's think step by step' after the dialogue history of the prompt. 
For the ToT method, we follow the official code to run it\footnote{ToT:\url{https://github.com/princeton-nlp/tree-of-thought-llm}
}. We run the four algorithm methods based on the GPT-4 model.

If the token number of the input prompt is beyond the token limit, we cut the PPT file content to reduce the token number of the prompt.
\subsection{Main results}
\begin{table*}[h]
\centering
\scalebox{0.6}{
\begin{tabular}{c|ccc|ccc|ccc|ccc}
\hline
\multirow{3}*{Models and Methods}&\multicolumn{6}{c}{Creating new PPT}&\multicolumn{6}{c}{Editing PPT template}\\
\cline{2-13}
{}&\multicolumn{3}{c|}{Turn-based}&\multicolumn{3}{c|}{Session-based}&\multicolumn{3}{c|}{Turn-based}&\multicolumn{3}{c}{Session-based}\\
\cline{2-13}
{}&\multicolumn{1}{c}{Accuracy}&\multicolumn{1}{c}{Avg token}&\multicolumn{1}{c|}{Avg API}&\multicolumn{1}{c}{Accuracy}&\multicolumn{1}{c}{Avg token}&\multicolumn{1}{c|}{Avg API}&\multicolumn{1}{c}{Accuracy}&\multicolumn{1}{c}{Avg token}&\multicolumn{1}{c|}{Avg API}&\multicolumn{1}{c}{Accuracy}&\multicolumn{1}{c}{Avg token}&\multicolumn{1}{c}{Avg API}\\
\hline
TD-003 & 72.6 & 2.8k & 3.0 & 12.7 & 20.8k & 23.9 & 24.4 & 2.9k & 8.1 & 4.0 & 13.2k & 26.6 \\
\hline
ChatGPT & 70.6 & 2.9k & 3.2 & 12.7 & 20.0k & 23.4 & 26.3 & 4.1k & 7.9 & 2.0 & 9.2k & 22.9 \\
\hline
GPT-4 & 75.1 & 2.9k & 2.9 & 22.7 & 20.8k & 22.4 & 38.1 & 7.5k & 7.8 & 6.0 & 24.1k & 24.7 \\
\hline
\hline
LLaMa-2 & 16.4 & 2.8k & 3.9 & 3.4 & 21.6k & 24.7  & 8.7 & 2.2k & 7.2 & 0.0 & 9.5k & 15.6 \\
\hline
Code-LLaMa & 36.8 & 2.8k & 3.4 & 0.0 & 20.7k & 32.1  & 18.7 & 3k & 7.3 & 2.0 & 9.6k & 22.6 \\
\hline
WizardLM & 23.9 & 1.3k & 3.3 & 4.3 & 12.5k & 22.4 & 10.0 & 1.3k & 5.7 & 0.0 & 4.3k & 16.5\\
\hline
Vicuna-v1.5 & 24.3 & 1.3k & 3.9 & 2.2 & 11.0k & 33.7 & 6.8 & 1.3k & 6.7 & 0.0 & 4.3k & 22.7\\
\hline
Baichuan & 15.5 & 1.3k & 9.8 & 0.0 & 10.9k & 44.7 & 4.4 & 1.3k & 9.6 & 0.0 & 4.3k & 24.3 \\
\hline
Baichuan-2 & 16.3 & 1.3k & 9.1 & 3.6& 11.6k & 48.9 & 8.7 & 1.3k & 9.2 & 0.0 & 4.2k & 22.3 \\
\hline
\end{tabular}
}
\caption{We report the results of LLMs in this table.' TD-003' is the Text-Davinci-003 model. We directly use the prompts in Figure~\ref{inference_prompt} to prompt LLMs to generate the API sequence.}
\label{tab.main}
\end{table*}

\begin{table*}[h]
\centering
\scalebox{0.58}{
\begin{tabular}{c|ccc|ccc|ccc|ccc}
\hline
\multirow{3}*{Models and Methods}&\multicolumn{6}{c}{Creating new PPT file}&\multicolumn{6}{c}{Editing PPT template}\\
\cline{2-13}
{}&\multicolumn{3}{c|}{Turn-based}&\multicolumn{3}{c|}{Session-based}&\multicolumn{3}{c|}{Turn-based}&\multicolumn{3}{c}{Session-based}\\
\cline{2-13}
{}&\multicolumn{1}{c}{Accuracy}&\multicolumn{1}{c}{Avg token}&\multicolumn{1}{c|}{Avg API}&\multicolumn{1}{c}{Accuracy}&\multicolumn{1}{c}{Avg token}&\multicolumn{1}{c|}{Avg API}&\multicolumn{1}{c}{Accuracy}&\multicolumn{1}{c}{Avg token}&\multicolumn{1}{c|}{Avg API}&\multicolumn{1}{c}{Accuracy}&\multicolumn{1}{c}{Avg token}&\multicolumn{1}{c}{Avg API}\\
\hline
GPT-4 & 75.1 & 2.9k & 2.9 & 22.7 & 20.8k & 22.4 & 38.1 & 7.5k & 7.8 & 6.0 & 24.1k & 24.7 \\
\hline
GPT-4+CoT & 77.0 & 2.9k & 3.1 & 23.1 & 20.8k & 22.7 & 40.6 & 7.5k & 8.0 & 6.0 & 24.1k & 25.2 \\
\hline
GPT-4+ToT & 76.5 & 20.8k & 3.0 & 21.8 & 146.4k & 22.6 & 40.6 & 81k & 7.6 & 4.0 & 256.8k & 24.0 \\
\hline
GPT-4+Content selection & 77.5 & 3.4k & 3.0 & 21.8 & 24.5k & 22.0 & 43.1 & 5.8k & 8.0 & 4.0 & 18.7k & 25.2 \\
\hline
GPT-4+API selection & 76.4 & 1.5k & 2.9 & 18.8 & 10.6k & 21.3 & 38.1 & 7k & 8.0 & 10.0 & 22.4k & 25.8 \\
\hline
\end{tabular}
}
\caption{We report the results of GPT-4 and algorithms based on the GPT-4 model. 'CoT' and 'ToT' are the chain of thought and tree of thought algorithms.}
\label{tab.algorithm}
\end{table*}
We report the results of LLMs in both turn-based and session-based evaluations in Table~\ref{tab.main} and ~\ref{tab.algorithm}.
From the results, we highlight the following key findings.
\par{\textbf{(1) Superior Performance of GPT-4:}} GPT-4 consistently outperforms other closed-source and open-source LLMs in both two tasks. Impressively, GPT-4 achieves 75.1\% turn-based accuracy in the creating new PPT file task, demonstrating its strong capability to finish one turn of the user instruction. GPT-4 also has a lower API cost compared to other closed-source LLMs since its precise API usage. GPT-4 incurs the highest token expense when editing PPT templates. That is because its higher token limit than other LLMs allows us to input more PPT template content.

\par{\textbf{(2) Code continual pre-training and further instruction finetuning can boost open-source LLMs' performance.}}: Based on Table~\ref{tab.main}, it's evident that current open-source LLMs struggle to match the performance of closed-source LLMs. For example, LLaMa-2-chat only achieves 16.2\% turn-based accuracy in the creating new PPT file task, which is far from the performance achieved by closed-source LLMs. 
We further find that code continual pretraining (Code-LLaMa) and instruction fine-tuning based on LLaMa-2 (WizardLM and Vicuna) can further improve LLaMa-2 performance obviously. For example, Code-LLaMa improves LLaMa-2's turn-based accuracy in the creating new PPT file task by 20.4 \%.  This observation suggests that there's untapped potential in open-source LLMs when it comes to our benchmark, and this potential can be unlocked further by code pre-training and enhancing instruction following ability. 
\par{\textbf{(3) Planning and selection algorithms can improve LLMs' turn-based performance}}
From Table~\ref{tab.algorithm}, we observe that the planning algorithms (CoT and ToT) can further improve the turn-based performance of GPT-4 by 1$\sim$2 percent. However, we surprisingly find that the more complex ToT algorithm does not outperform the zero-shot CoT algorithm with a 5$\sim$10 times token cost.  Content and API selection algorithms can further improve the turn-based performance of GPT-4 by 1$\sim$ 5 percent. That is because they reduce the task difficulty by filtering irrelevant PPT content/APIs in the input prompt. The API selection algorithm also reduces the average token cost by reducing the number of APIs listed in the prompt. However, for the challenging session-based evaluation, these algorithms can not improve GPT-4's performance or improve it slightly.

\subsection{Three challenges in our PPTC benchmark}
\begin{figure*}[t]
  \centering
\includegraphics[height=0.315\textwidth, width=0.995\textwidth]{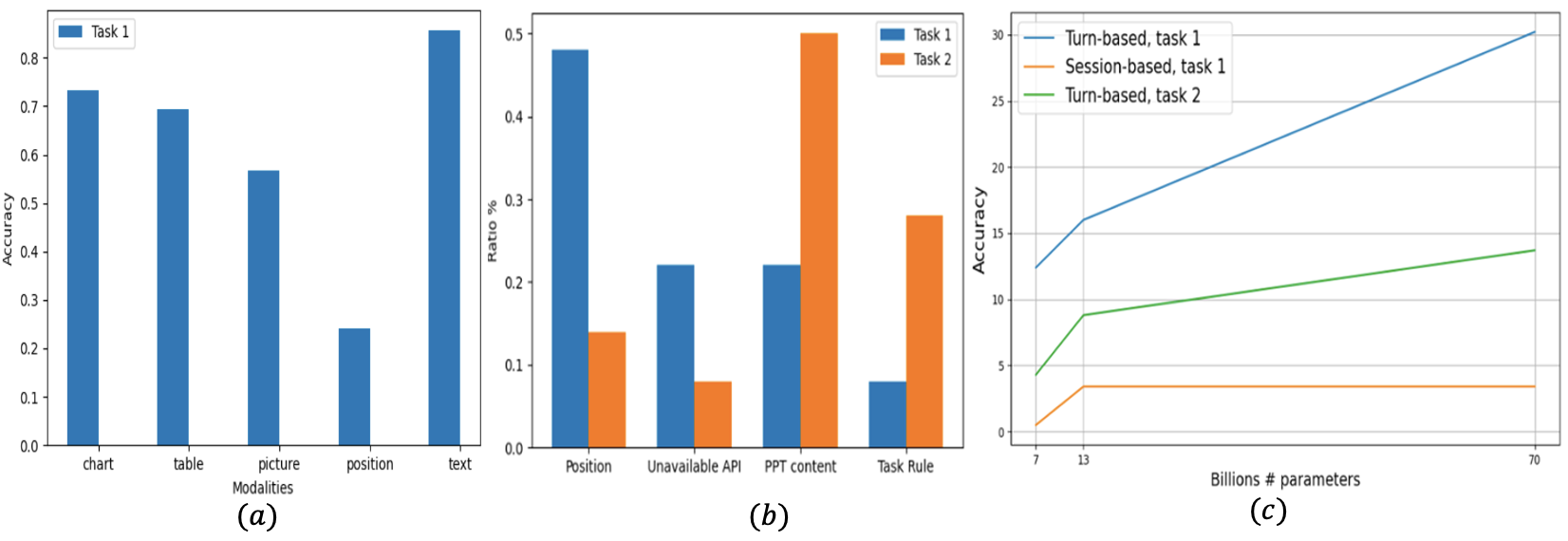}
  \caption{We illustrate the analysis results of the creating new PPT file task (task 1) and the editing PPT template task (task 2). In sub-figure (a), we report the average turn-based accuracy for instructions involving chart, table, picture, position, and pure text. We don't draw the accuracy of task 2 as no chart instruction in this task. In sub-figure (b), we report the ratio of four common errors made by GPT-4. In sub-figure (c), we report the accuracy with the model size. We don't plot the session-based accuracy of task 2 as it is zero.}
  \label{fig:analysis}
\vspace{-10pt}
\end{figure*}

From the result Table~\ref{tab.main} and Figure~\ref{fig:analysis}. we highlight the following three key challenges.
\par{\textbf{(1) Error accumulation makes LLMs performance poor in finishing the entire multi-turn session.}}: The performance of all LLMs in handling sessions consisting of multiple turns is notably poor. Even GPT-4, which performs well in turn-based evaluation, achieves only a 22.7\% session-based accuracy for the "creating new PPT file" task and a mere 6.0\% session-based accuracy for the "editing PPT template" task. Current planning algorithms usually fail to improve session-based accuracy. In some cases, they can even make the performance worse. 
The session-based evaluation is challenging since errors made in previous turns make the LLM fail to finish the session and also influence the completion of the current turn. Also, we need more advanced planning algorithms to complete the multi-turn session.
\par{\textbf{(2) LLMs perform badly in processing long PPT template:}} Current LLMs' performance in the editing PPT temples task is pretty poor. For example, the strongest GPT-4 only achieves 38.1\% turn-based accuracy and 6.0\% session-based accuracy in this task. Other LLMs' performance is even poorer. The content selection algorithm can partially solve this challenge by filtering out irrelevant file content, but GPT-4 with it still only achieves 43.1\% turn-based accuracy. That means current LLMs (e.g., GPT-4) still struggle to handle complex and lengthy PPT templates. For open-source LLMs, there's a risk of information loss due to token limitations (typically 2$\sim$4K tokens limit), which often require truncating lengthy PPT content. When it comes to session-based performance, the accuracy remains nearly zero. This implies that current LLMs are still far from being ideal PPT agents capable of effectively assisting users in editing PPT templates during a multi-turn dialogue session.

\par{\textbf{(3) Multi-modal instructions increase the LLM's failure rate significantly.}
To assess LLMs' task completion performance for instructions involving multi-modal operations (Table, Chart, Picture, Position, and text), we calculate the average accuracy of GPT-4 for instructions involving each modality, respectively. This is done by dividing the number of correctly completed instructions within each modality by the total number of instructions involving that modality's operation. The results are presented in Figure~\ref{fig:analysis} (a). From the figure, we observe that GPT-4 performs exceptionally well in the text modality, achieving an accuracy of 85.6\%. Its performance becomes poorer when processing structured data (Chart and Table), with 12.4\% and 16.2\% lower accuracy. Instructions involving picture-related operation pose an even greater challenge for GPT-4, as it achieves a 56.8\% turn-based accuracy in this modality. GPT-4 exhibits its weakest performance in instructions involving position-related (spatial) operations, with only 24\% accuracy. This underscores GPT-4's limitations in spatial perception ability. 

\section{Analysis}
In this section, we analyze the reasons for GPT-4's errors. We further analyze the influence of model size and dialogue history.

\subsection{\textit{Error Analysis of GPT-4 in our benchmark}}
To analyze the error made by GPT-4, in our benchmark, we gather 50 wrong samples for each of the two tasks in our benchmark in the turn-based evaluation.
We find that these wrong samples fall into four error types and visualize the distribution of these four main error types in Figure~\ref{fig:analysis} (b): (1) Position errors: These occur when GPT-4 struggles with instructions involving position adjustments. For example, when asked to move the shape to the bottom of the slide, GPT-4 wrongly calls the "set$\_$top" API. position error is the main error in the creating new PPT file task as this task contains many instructions involving position operation. (2) Calling unavailable APIs: GPT-4 sometimes generates APIs that don't actually exist in the reference API file, resulting in what we call the "API hallucination problem." (3) Misunderstanding PPT file content: GPT-4's comprehension of the PPT content can be flawed, leading to the generation of incorrect API sequences. For example, when instructed to make the font size of the current slide's title consistent with previous slides, GPT-4 set a font size that is different from what was used in previous slides' titles. In the editing template task, misunderstanding the PPT content becomes the main error since this task needs to understand the complex PPT template. (4) Unfollowing Powerpoint task rules: Completing Powerpoint tasks demands a deep understanding of various task rules. For instance, writing a new slide title requires first deleting the original text and then inserting the new title into the text box. However, GPT-4 may directly insert the new content.  
For the session-based evaluation, we also collect 50 wrong examples. We find that the main reason for the poor session-based performance is the LLM fails to finish the session once it makes an error in one turn of the session. The reasons for errors made in a single turn are similar to those in the turn-based evaluation. One unique phenomenon in this evaluation is that the LLM would repeat previous errors (e.g., employing infeasible APIs) in subsequent turns. 
\subsection{\textit{Does bigger LLM work better on PPTC?}}
To investigate how the model size impacts the LLM's performance in our benchmark, we conduct tests using LLaMa-2-chat LLM with 7, 13, and 70 billion parameters and plot the results in Figure~\ref{fig:analysis} (c). We observe that larger LLM consistently achieve higher turn-based accuracy for both the creating new PPT and editing PPT template tasks. For example, in the creating new PPT file task, we find that the turn-based accuracy increases from 13.2 (7B) to 30.1 (70B). However, we do not observe a clear positive correlation between model size and session-based performance. One possible explanation is that although the 70B LLM can correctly finish more intermediate steps, it still falls short of completing the entire session. To improve the session-based performance, a larger LLM may be necessary.
\subsection{\textit{Does dialogue history help LLMs to generate the API sequence?}}
To investigate the influence of dialogue history in our prompt (see Figure~\ref{inference_prompt}), we make an ablation experiment for the dialogue history component of our turn-based evaluation prompt\footnote{The task instruction, current user instruction, API file, PPT content in the prompt are necessary parts for generating the API sequence. So we don't conduct ablation studies on them.}. In this evaluation, the dialogue history contains previous turns along with their feasible API sequences. When we removed the dialogue history from the prompt, we observed a decline in GPT-4's performance. 
Specifically, GPT-4 drops its performance from 75.1 \% to 73.1 \% in the creating new PPT file task and decreases its performance by 6.2 \% in the editing template task. This experiment shows the positive effect of the dialogue history, as it helps the LLM to both understand the dialogue background and instruct the LLM to correctly use the APIs, similar to few-shot demonstration examples.  

\section{Related Works}
\textbf{Large Language Models} like ChatGPT, GPT-4~\cite{bubeck2023sparks,openai2023gpt4}, and Bard have billions of parameters and have been trained on the Internet corpus with trillions of tokens. They can write code~\cite{liu2023your}, prove mathematical theorems~\cite{jiang2022draft}, pass the professional exam~\cite{zhong2023agieval,gilson2023does,katz2023gpt}, and also perform well on other basic natural language tasks~\cite{kim2023language,jiao2023chatgpt,zhong2023agieval,wang2023chatgpt}. That raises the hope of achieving artificial general intelligence (AGI). 

To further boost LLM's performance on the specific task, one approach involves prompting engineerings, such as the chain of thought prompting~\cite{wei2022chain,shi2022language,yao2023tree}, self-consistency~\cite{wang2022self} and the least to most prompting~\cite{zhou2022least}. Another approach aims to use feedback to improve performance. The self-refine method~\cite{madaan2023self} refines the output through iterative feedback and refinement Provided by LLM itself. The Reflexion~\cite{shinn2023reflexion} method generates and stores the reflection based on the sparse reward signal and then uses the reflection to induce better decisions in subsequent trials. The learning to program method~\cite{guo2023learning} learns the task program by inducing the general solutions from the errors (feedback) iteratively and uses the program to guide the inference. 

\textbf{Task completion benchmarks for measuring large language models}. To measure LLM's task completion performance,  
Saycan~\cite{brohan2023can} and VirtualHome~\cite{puig2018virtualhome} benchmarks ask LLM to generate the correct action sequence for controlling the robot to finish user instruction. WebShop~\cite{yao2022webshop} and  Android in the wild~\cite{rawles2023android} ask LLM to navigate websites and conduct actions to meet the user requirement. APIBench~\cite{patil2023gorilla} and ToolBench~\cite{xu2023tool,qin2023tool} involve selecting and using APIs to complete the task instruction. Agentbench~\cite{liu2023agentbench}assesses LLM as autonomous agents in 8 environments and WebArena~\cite{zhou2023webarena} considers task completion in web-based interactions. 

\textbf{AI assistant system for complex task completion} For more complex tasks that involve using tools and utilizing environmental information, there are many strong AI systems (e.g., TaskMatrix~\cite{liang2023taskmatrix}) that help the user finish the complex task. One approach involves connecting massive models and tools with the Large Language Model for task completion. Examples include Visual ChatGPT~\cite{wu2023visual} and HuggingGPT~\cite{shen2023hugginggpt} which use LLM to deploy task-specific models to finish the user instruction based on the observation of task information (e.g., visual information), Voyager~\cite{wang2023voyager} that uses the fixed LLM to continually learn skills (tools) based on the observation of the Minecraft environment. 
Another approach involves training an end-to-end LLM to finish the user instruction. Examples include Gorilla~\cite{patil2023gorilla} for generating API calls to finish the user query by using the API bench to fine-tune the pre-trained LLaMA~\cite{touvron2023llama} model. PaLM-E~\cite{driess2023palm} for various robot reasoning tasks by fine-tuning the pretrained PaLM~\cite{chowdhery2022palm} model using features from sensor modalities. Different from the above systems, we focus on the research topic of the AI assistant system in office software. 
\section{Conclusion}
We introduce the PowerPoint Task Completion benchmark to measure LLMs' ability to complete user instructions within the context of the PowerPoint software. It contains hundreds of multi-turn sessions with different topics and thousands of instructions with varying levels of difficulty. We further propose the PPTX-evaluation system to access and compare the performance of different LLMs. Results show that GPT-4 is the strongest LLM but still performs poorly in finishing entire sessions. We further analyze the behavior of LLMs and find three main error factors that limit their performance. Our benchmark and findings can help the research community design better AI task completion assistants.

\section{Limitations}
Our benchmark does not consider instructions that involve subjective evaluation. For example, the user may want to make the slide more beautiful. However, it's hard to automatically evaluate if the generated file (the model output) is more beautiful. Another limitation is that we do not consider the instructions that need non-API operations. For example, the user may want to draw a cat on the slide. That instruction needs the AI-assistant system to draw the cat by dragging the mouse and is still infeasible for LLMs and LLM-based systems. We only consider instructions that can be completed by directly executing the API sequence. 
\bibliography{anthology}

\appendix
\section{The API Reference File}
We list all APIs and their descriptions in  Figures~\ref{fig.apis1} and ~\ref{fig.apis2}. We provide 49 feasible APIs.
\begin{figure*}
\tcbset{colback=green!-15!white, colframe=green!65! black, fonttitle=\bfseries}
\begin{tcolorbox}[title=API reference file, sidebyside align=top]
\textbf{Slide-related APIs}

API: create slide(): This API creates a new slide.

API: move to previous slide(): This API moves to the previous slide.

API: move to next slide(): This API moves to the next slide.

API: move to slide(slide id): This API moves to the slide with given slide id.It takes one parameter 'slide id', the ID of the slide to move to as a integer.

\textbf{Choose-related APIs}

API: choose title(): This API selects the title on the slide. You should first call choose title() before inserting text to or changing font attributes of the title. 

API: choose content(): This API select the content on the slide. You should first call choose content() before inserting text to or changing font attributes of the content. 

API: choose textbox(idx): This API selects the textbox element on the slide. It takes one parameter, the index of textbox as integer. idx is set to 0 by default, meaning the first textbox. You should first call choose textbox() before inserting text to or changing font attributes of the textbox element. 

API: choose picture(idx): This API selects the picture element on the slide. It takes one parameter, the index of textbox as integer. idx is set to 0 by default, meaning the first textbox. You should first call choose picture() before changing height, width, rotation of the picture element. You should not call choose picture() before inserting picture element. 

API: choose chart(): This API selects the chart element on the slide. You should first call choose chart() before changing the chart. You should not call choose chart() before inserting chart element. 

API: choose shape(shape name): This API selects a specific shape by shape name on the slide. It takes one parameter 'shape name', the name of the shape to select as a string.         shape name can be chosen from ['rectangle','right arrow','rounded rectangle','triangle','callout','cloud','star','circle'] You should first call choose shape(shape name) before you can do operations on the shape. You should not call choose shape(shape name) before inserting shape element. 

API: choose table(): This API selects the table element on the slide. You should first call choose table() before changing the table. You should not call choose table() before inserting table element.

API: choose table cell(row id, column id): This API selects a specific cell in the table by giving row id and column id. It takes two parameters, the row id and column id of the cell to select as integers (id starts from 0). Remember the first parameter is row id, the second parameter is column id. You should first call choose table cell(row id, column id) before inserting text. 

\textbf{Basic APIs}

API: set background color(color): This API sets the background color of the slide. It takes one parameter 'color', the color name to set as a string, such as 'red', 'purple'. 

API: set width(width): This API sets the width of the selected object. It takes one parameter 'width', the width of an object in centimeters as float. You should first choose an object before you can change the width of it. 

API: set height(height): This API sets the height of the selected object. It takes one parameter 'height', the height of an object in centimeters as float. You should first choose an object before you can change the height of it 

API: rotate element(angle): This API rotates the selected element by the specified angle. It takes one parameter 'angle', the angle to rotate clockwise as integer. You should first choose an object before you can rotate it. 

API: set fill color(color): This API sets the fill color of the selected object after the object is chosen. It takes one parameter 'color', the color name to set as a string, such as 'red', 'purple'. You can set the fill color of content, title or textbox. 

API: set left(left): This API moves and changes the object's position. It sets the x position of the selected object's leftmost point. It takes one parameter, the x position to set. You should first choose an object before you can change the left of it

API: set top(top): This API moves and changes the object's position. It sets the y position of the selected object's upmost point. It takes one parameter, the y position to set. You should first choose an object before you can change the top of it.
\end{tcolorbox}
\captionof{figure}{The reference API file: part 1.}
\label{fig.apis1}
\vspace{-5pt}
\end{figure*}

\begin{figure*}
\tcbset{colback=green!-15!white, colframe=green!65! black, fonttitle=\bfseries}
\begin{tcolorbox}[title=API reference file, sidebyside align=top]
\textbf{Text-related APIs}

API: insert text(text): This API inserts text into a text frame (textbox, title, content, table). 

API: insert bullet point(text): This API inserts a bullet point into the content. It takes one parameter, the text of the bullet point to insert as a string. 

API: insert note(text): This API inserts a note onto the slide. It takes one parameter, the note text to insert as a string. 

API: insert textbox(): This API inserts a textbox onto the slide. When you need to add a caption or text under/above/left to/right to an object, you can call insert textbox(). 

API: delete text(): This API delete the text part of an object. You should first choose content or title before you can call delete text() 

API: set font size(font size): This API sets the size of the font It can take one argument 'font size', the font size to set as an integer. 

API: set font color(color): This API sets the color of the font. It takes one parameter 'color', the color name to set as a string, such as 'red', 'purple'. 

API: set font bold(): This API sets the font to be bold. 

API: set font italic(): This API sets the font to be italic. 

API: set font underline(): This API sets the font to be underlined. 

API: set font style(font name): This API sets the font style of the selected text. It can take one argument 'font style', the font name as a string. 

API: set line space(line space level): This API sets the line spacing of the selected text. It can take one argument 'line space level', as an integer, default 0. 

API: text align left(): This API aligns the text to left. 

API: text align center(): This API aligns the text to center. 

API: text align right(): This API aligns the text to right. 

\textbf{Image and shape-related APIs}

API: insert picture(picture name): This API inserts a picture onto the slide. It takes one parameter 'picture name', the name or description of picture as a string 

API: insert rectangle(): This API inserts a rectangle or square shape onto the slide. 

API: insert right arrow(): This API inserts an arrow shape onto the slide. 

API: insert rounded rectangle(): This API inserts a rounded rectangle shape onto the slide. 

API: insert triangle(): This API inserts a triangle shape onto the slide. 

API: insert callout(): This API inserts a callout shape onto the slide. 

API: insert cloud(): This API inserts a cloud shape onto the slide. 

API: insert star(): This API inserts a star shape onto the current slide. 

API: insert circle(): This API inserts a circle or oval shape into the current slide. 

\textbf{Table-related APIs}

API: insert table(row num, col num): This API inserts a table of row num rows and col num columns onto the current slide. It takes two argument, the row number and the column number of the inserted table as integer. Remember the first parameter is row number and the second parameter is column number. 

API: insert table row(row data): This API inserts a row (list) of data into the table. It takes one argument, the data to insert as a list of numbers or strings. You should first call choose table() before you can call insert table row(). The parameter 'row data' should be a list of strings. 

\textbf{Chart-related APIs}

API: insert line chart(data, series): This API inserts a line chart onto the slide. It takes two argument, 'data' is a list of numbers and 'series' is a list of strings. 

API: insert bar chart(data, series): This API inserts a bar chart onto the slide. It takes two argument, 'data' is a list of numbers and 'series' is a list of strings. 

API: insert pie chart(data, series): This API inserts a pie chart onto the slide. It takes two argument, 'data' is a list of numbers and 'series' is a list of strings. 

API: set chart title(title): This API sets the title of a previously inserted chart. It takes one argument 'title', the title to be set as a string.
\end{tcolorbox}
\captionof{figure}{The reference API file: part 2.}
\label{fig.apis2}
\vspace{-5pt}
\end{figure*}
\label{appendix:apis}
\section{The Prompt for Content Selection Algorithm}
We put the prompt of content selection algorithm in Figure~\ref{fig.content_selection}.
\begin{figure*}
\tcbset{colback=green!-15!white, colframe=green!65! black, fonttitle=\bfseries}
\begin{tcolorbox}[title=Content Selection prompt, sidebyside align=top]
\small{You are an AI assistant for PowerPoint. Your task is to determine what kind of content is necessary to fulfill the user's instruction.
You have an API to extract the content, please call the get$\_$content api with correct parameters to fulfill the user's instruction.
You need to extract the minimum necessary information to fulfill user's instruction.
\\\\
\textbf{Get$\_$content\quad API:} get$\_$content(need$\_$text: Indicates whether text information is required. The text information encompasses text in title, content, textbox, table, chart, and shape. This parameter is particularly useful when inserting or modifying text of title, content, textbox, table, chart, and shape, or when information about these objects is essential.
    
    need$\_$style: Indicates whether style information is required. Style information includes attributes like font type, font size, color, background color, line space, bold, undeline, italic and other visual aspects of objects like rotation. This is useful when changing the appearance of text or objects or when information about an object's appearance is essential.
    
    need$\_$position: Indicates whether position information is required. The position details encompass an object's height, width, and its left and top positions. This is crucial when moving objects or altering an object's size.
    
    need$\_$title: Determines if information related to the title is required.
    
    need$\_$content: Determines if information related to the content is required.
    
    need$\_$picture: Determines if information related to the picture is required.
    
    need$\_$table: Determines if information related to the table is required.
    
    need$\_$chart: Determines if information related to the chart is required.
    
    need$\_$textbox: Determines if information related to the textbox is required.
    
    need$\_$shape: Determines if information related to the shapes (rectangle, right arrow, rounded rectangle, triangle, callout, cloud, star, circle) is required.
)

Where the parameters are either 1 (needed) or 0 (not needed).
You should only answer with calling get$\_$content() with the right parameters.
\\\\
\textbf{For examples:}

Instruction: 
Increase the font size of the content to 20. 

Explanation: 
For information, style information (font size) is needed. 
For objects, content is needed.

Answer:

get$\_$content(need$\_$text=1,need$\_$style=1,need$\_$position=0,\\need$\_$title=0,need$\_$content=1,need$\_$picture=0,need$\_$\\table=0,need$\_$chart=0,need$\_$textbox=0,need$\_$shape=0)

\textbf{...}

\textbf{Given the instruction, output the Answer without Explanation:}

Instruction:
<Current user instruction>

Answer:}

\end{tcolorbox}
\captionof{figure}{The prompt of the content selection algorithm.}
\label{fig.content_selection}
\vspace{-5pt}
\end{figure*}
\label{appendix:content_selection}
\end{document}